\documentclass[10pt,twocolumn,letterpaper]{article}

\usepackage{wacv}
\usepackage{times}
\usepackage{epsfig}
\usepackage{graphicx}
\usepackage{amsmath}
\usepackage{amssymb}

\usepackage{multirow}
\usepackage{makecell}
\usepackage{wrapfig}
\usepackage{graphicx}
\usepackage{lipsum}
\usepackage{kotex}
\usepackage{enumitem}
\usepackage{amsmath}
\usepackage{amssymb}
\usepackage{comment}
\usepackage{float}
\usepackage{color, colortbl}
\usepackage{authblk}
\usepackage[para]{footmisc}

\usepackage[ruled,vlined]{algorithm2e}
\include{pythonlisting}

\definecolor{OliveGreen}{rgb}{0.5, 0.5, 0.0}
\definecolor{Gray}{gray}{0.9}

\newcommand{\nj}[1]{\textcolor{black}{#1}}
\newcommand{\hj}[1]{\textcolor{black}{#1}}

\def\wacvPaperID{834} 

\wacvfinalcopy 

\ifwacvfinal
\def\assignedStartPage{1} 
\fi

\ifwacvfinal
\usepackage[breaklinks=true,bookmarks=false]{hyperref}
\else
\usepackage[pagebackref=true,breaklinks=true,colorlinks,bookmarks=false]{hyperref}
\fi

\ifwacvfinal
\else
\fi

\begin{document}

\title{Few-Shot Object Detection  by Attending to Per-Sample-Prototype}


\author[ ]{Hojun Lee$^1$\thanks{This work is conducted during the author’s research internship at NAVER WEBTOON Corp. \hspace{-0.5cm}} \:\: Myunggi Lee$^{1,2}$ \:\: Nojun Kwak$^1$\thanks{Corresponding author}}
\affil[ ]{Seoul National University$^1$  \: NAVER WEBTOON$^2$ }
\affil[ ]{\tt\small \{hojun815, nojunk\}@snu.ac.kr  \: myunggi@webtoonscorp.com}

\maketitle

\begin{abstract} \vspace{-0.2cm}
Few-shot object detection aims to detect instances of specific categories in a query image with only a handful of support samples. Although this takes less effort than obtaining enough annotated images for supervised object detection, it results in a far inferior performance compared to the conventional object detection methods. In this paper, we propose a meta-learning-based approach that considers the unique characteristics of each support sample. 
Rather than simply averaging the information of the support samples to generate a single prototype per category, our method can better utilize the information of each support sample by treating each support sample as an individual prototype.
Specifically, we introduce two types of attention mechanisms for aggregating the query and support feature maps. The first is to refine the information of few-shot samples by extracting \nj{shared} information between the support samples through attention. Second, each support sample is used as a class code to leverage the information by comparing similarities between each support feature and query features. Our proposed method is complementary to the previous methods, making it easy to plug and play for further improvement. We have evaluated our method on PASCAL VOC and COCO benchmarks, and the results verify the effectiveness of our method. In particular, the advantages of our method \hj{are} maximized when there is more diversity among support data.
\end{abstract}\vspace{-0.4cm}

\section{Introduction} \vspace{-0.05cm}

Multi-object detection is a classical computer vision task of recognizing and localizing the instances of specific objects categories from a given scene. 
In virtue of abundant images with bounding box annotations, object detection has experienced an enormous advancement with numerous deep learning-based approaches \cite{ren2015faster, liu2016ssd, redmon2016you}.
Notwithstanding its remarkable achievements, the methods still have difficulty \nj{in} learning novel object categories when the number of labeled data samples is \nj{small} \cite{sohn2020simple, li2020overcoming}. 
Few-shot learning problems address such issues, 
which is common in real-world cases. 
However, learning few-shot samples by empirical risk minimization in a supervised manner easily overfits and may result in poor generalization \cite{bousquet2003introduction, wang2020generalizing}.

To alleviate this problem, several approaches have been studied, and \textit{meta-learning} is one of the most successful ones in the few-shot classification scenario. In the few-shot setting, the model is given a small number of labeled \emph{support} data for training, and at the time of inference, an input \emph{query} image is classified as one of the support categories. Metric-based classification frameworks \cite{vinyals2016matching, yang2021free, snell2017prototypical, li2019few, sung2018learning}, \nj{one of the popular meta-learning methods,} firstly calculate the centroid of each support class called \emph{class prototype} from the support data, and then classify \nj{the query} by measuring the similarity of the prototype with the query.
\begin{figure*}
\begin{center}
\includegraphics[width=17.5cm]{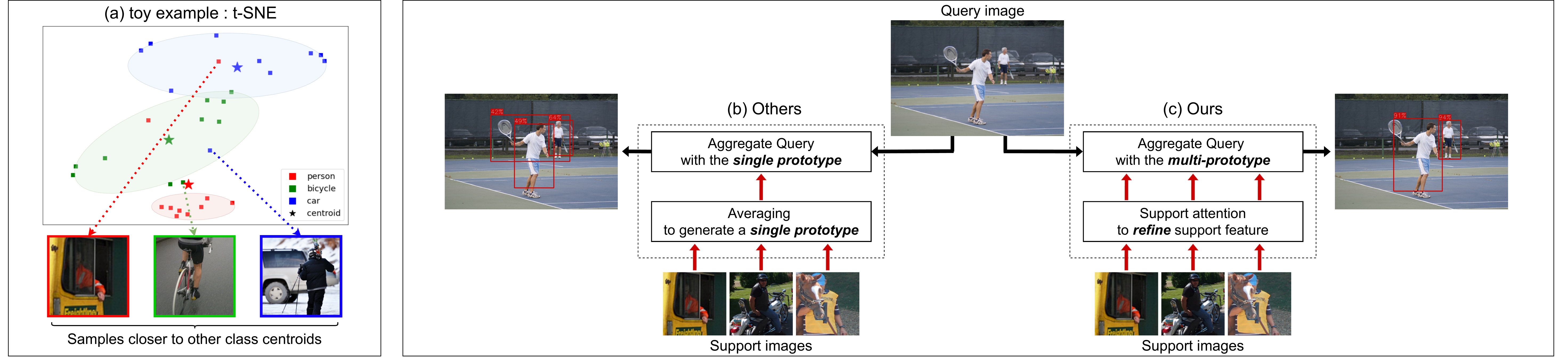}
\vspace{-0.70cm}
\end{center}
  \caption{\textbf{\nj{Concept of our method:}} LEFT: (a) Toy example when support data \nj{have large diversity and are} misclustered. RIGHT: (b) Feature vectors of collected $K$-shot support images of a \textit{person} contain diverse information and just averaging this information for aggregation with the query may deteriorate the detection performance. (c) Instead of using just a single averaged prototype per class, we use one prototype for each support image which has been refined by attending to other support images.}  \vspace{-0.20cm}
\label{fig:intro}
\end{figure*}

\nj{Since a significant progress has been made in the literature of few-shot classification problems, the problem of few-shot learning for object detection (FSOD) has also been studied. One of the successful approaches for FSOD is to extend typical meta-learning approaches to FSOD.}
One of the key issues in this \nj{line of} research is how to aggregate the class prototype with the query image \cite{xiao2020few, kang2019few, fan2020fsod, yan2019meta, wang2019meta}.
Although there have been performance improvements of FSOD through previous methods based on meta-learning, previous aggregation methods have \nj{a couple of} main problems.

First, a handful of support data may be noisy and this can cause unexpected \nj{side effects}.
For example, instances of different categories may be \nj{close to} one another or instances of the same class may differ in shape and perspective\hj{, which causes some samples  to be far from intra-class samples in the feature space (Fig. \ref{fig:intro} (a))}.
Therefore, if the information is not refined before averaging the support data, since it is a few-shot, the \nj{averaged prototype} may be far from the centroid of the real distribution.
Second, to our best knowledge, every method studied so far relies on class-wise \textit{single representative} by averaging the information of the support data, which \nj{are compared} with the query data \hj{(Fig. \ref{fig:intro} (b))}. Instances in query images have large variations in size, perspective and even a possibility of occlusion. Furthermore, both the query and support images may have multiple \nj{instances with different categories} close to each other. Therefore, rather than 
generating a single prototype per category that covers all the diversity and abolishing other information, it may be more advantageous to make better use of the information contained in the support data.

\nj{To resolve these problems, we} propose a novel method (Fig. \ref{fig:intro} \hj{(c)}) consisting of two \nj{modules} that aggregate the query and support data. First, we propose a method to refine the support information through an attention mechanism among support data before aggregating the query and support data. Second, rather than averaging the information of the support image, we use each support image as a prototype, which we call \textit{per-sample prototype}. Through this method, we can better aggregate the diverse information of support data with queries. 

We have applied the proposed method to two \nj{different} architectures \cite{xiao2020few, fan2020fsod} in different ways. Our method improves the average precision (AP) for new unseen classes on PASCAL VOC  \cite{pascal-voc-2007, pascal-voc-2012} and COCO \cite{lin2015microsoft} benchmarks in both architectures. We also qualitatively verify that our method enhances the quality of the clusters available from support feature vectors of the same class via t-SNE \cite{van2008visualizing}.

Our contributions can be summarized as follows: \vspace{-0.01cm}
\begin{itemize}[leftmargin=+.15in] 
\vspace{-0.05cm}
    \setlength\itemsep{-0.3em} 
    \item We investigate that refinements of \hj{the support feature maps}
    induce useful information for \hj{aggregation} 
    through an attention mechanism.
    \item We propose a method that aggregates query and support features without using one prototype \hj{per class}, which allows fully leveraging information of support data.
    \item We show that our method can be applied to various types of architectures, and yields meaningful performance improvement on the  PASCAL VOC and COCO benchmarks compared to our baselines \cite{fan2020fsod, xiao2020few}. 
    \item \hj{Through t-SNE and clustering experiments, we demonstrate that intra-class support features are well clustered by our method, learning robust classifiable features.}
\end{itemize}

\section{Related Work} \vspace{-0.1cm}
\label{sec:related}
\noindent {}{\bf Object Detection}\quad It is the task of detecting instances of a specific category in an image. There have been many studies \cite{ren2015faster, redmon2017yolo9000, liu2016ssd} on supervised learning with large annotated image datasets. Also, several variant tasks have been studied. For example, \textit{weakly supervised object detection} \cite{kim2020tell, durand2017wildcat} is the task of learning to detect only with weak annotations (e.g., image-level category) without bounding box annotation. \textit{Semi-supervised object detection} \cite{jeong2019consistency, sohn2020simple} is a task using both labeled and unlabeled data, and \textit{few-shot object detection}, which we deal with in this paper, aims to detect instances of novel categories with few samples. \\
\indent Object detectors are largely divided into single-stage and two-stage detectors. The single-stage detectors \cite{liu2016ssd, lin2017focal} predict the object's class and bounding box directly from the features from the feature extractor. 
The two-stage detectors \cite{ren2015faster, cai2018cascade} detect objects in two steps: first, they generate class-agnostic candidate boxes using RPN (Region Proposal Network). Then, the candidate boxes are classified and the corresponding bounding boxes are regressed. In FSOD, both single-stage \cite{chen2018lstd, kang2019few} and two-stage methods \cite{yan2019meta, wang2020frustratingly, fan2020fsod, xiao2020few} have been studied. Following the majority trend, we devised our method that can be applied to two-stage detectors \nj{such as} Faster R-CNN \cite{ren2015faster}. 

\vspace{+0.08cm}
\noindent {\bf Meta-Learning}\quad Briefly speaking, it is a research topic to learn how to learn. There are several approaches of meta-learning, such as 1) gradient-based methods \cite{finn2017model, li2018learning} that learn to well-transfer the knowledge learned by several tasks to a new task and 2) model-based methods \cite{munkhdalai2017meta, ravi2016optimization} that aim to design structures that can generalize well, or utilize an external meta-learner or memory. 3) metric-based methods \cite{snell2017prototypical, vinyals2016matching, ren2018meta, yang2021free} that perform non-parametric learning by comparing query \nj{sample} with support \nj{samples} and predicting the category of test data by comparing with training support \nj{samples}. A typical strategy of applying metric-based learning to few-shot learning is to generate a single class prototype for each category by averaging features of support data belonging to the category \cite{snell2017prototypical, vinyals2016matching, sung2018learning}. 
If networks learn sampled mini-batches called \emph{episode} through meta-learning, \nj{an episode will consist} of sub-classes at each iteration. \nj{This episodic training strategy} has been shown to generalize better on novel few-shot data because \nj{it naturally} mimics the few-shot task. In this paper, instead of generating \nj{one prototype for each class}, we propose an alternative method, generating per-sample prototypes, that makes better use of the information of the support data by treating each sample as a prototype.

\vspace{+0.15cm}
\noindent{\bf Few-Shot Object Detection}\quad One of the most successful approaches in FSOD are meta-learning-based methods. Given a support data $S$ composed of $K$ samples of a specific category and a query image $Q$, the goal of FSOD methods based on meta-learning is to recognize and localize the instances of $Q$ with the help of $S$. MetaDet \cite{wang2019meta}, MetaYOLO \cite{kang2019few} and Meta R-CNN \cite{yan2019meta} proposed methods for meta-learner to generate a prototype per category from the support data and aggregate these prototypes with the query features by channel-wise multiplication. FsDetView \cite{xiao2020few} showed that when query features are aggregated with support features, it is more effective to concatenate three features together: \nj{the} channel-wise multiplication feature map of query and support, \nj{the} subtraction feature map of query and support, and \nj{the} query feature map itself. FewX \cite{fan2020fsod} proposed a method of aggregating query features and support features before the region proposal process, unlike the previous methods that aggregate after the region proposal process. All of these methods generate a single prototype per category when performing aggregation, but we propose a novel method of aggregating each sample \nj{through the per-sample prototype.}

\section{Approach}
\label{sec:approach}
\subsection{Problem Formulation}
\label{sec:problem formulation}
In the few-shot object detection scenario, we assume we have two sets of data sources, $D_{b}$ and $D_{n}$. $D_{b}$ is a \textit{base} dataset with abundant annotated instances of base classes $C_{b}$, and $D_{n}$ is a \textit{novel} dataset with few labeled instances of novel classes $C_{n}$. Here, \nj{we assume} there are no overlapping classes between $C_{b}$ and $C_{n}$, i.e., $C_{b} \cap C_{n} = \emptyset$. Few-shot object detection aims to train a detector with limited data source $D_{n}$ to recognize and localize novel instances of categories $C_{n}$ with the help of knowledge learned from the base dataset $D_{b}$.
In this paper, as with the previous researches \cite{li2019few, kang2019few, xiao2020few, wang2020frustratingly}, it is assumed that the novel dataset, $D_{n}$, is composed of $K$ annotated instances per category. Also, the number of novel categories is $N$, and we call this problem a $K$-shot, $N$-way few-shot object detection problem.

\subsection{Overall Architecture}
\label{subsec:overall}

Our method aims to find the novel instances in the query image by leveraging meta-learning. In the meta-learning scenario, we concentrate on two aspects of the pipeline: how to extract informative representation from support set $S$ and how to combine it with query features. To this end, we propose \nj{the} \textit{Intra-Support Attention Module} (ISAM) and \nj{the} \textit{Query-Support Attention Module} (QSAM), which \nj{can} complement various frameworks \nj{in a plug-and-play manner} and bring significant improvements in performance. 

\begin{figure*}
\centering
\includegraphics[width=16.5cm]{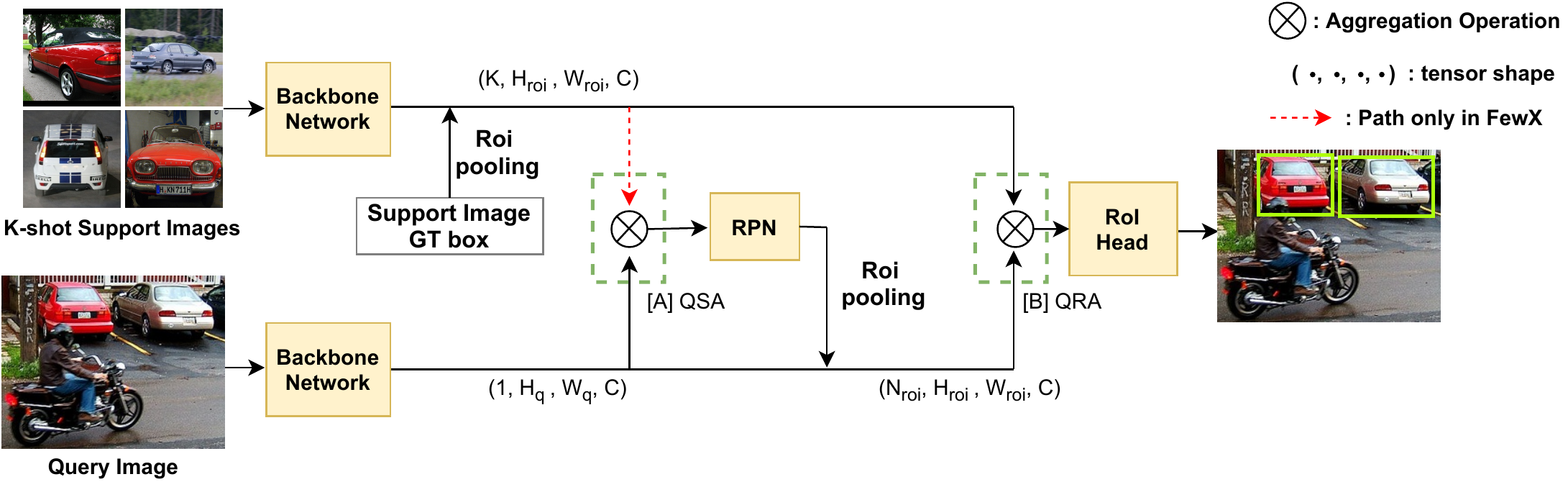}
\vspace{-0.1cm}
\caption{\textbf{\nj{Our} overall Architecture} based on Faster R-CNN \cite{ren2015faster} to find instances of support category in the query image. [A] and [B] are aggregation procedure (Fig. \ref{fig:aggregation}) for the query feature and the support features. Baselines are FewX \cite{fan2020fsod} and FsDetView \cite{xiao2020few}, where FewX has both [A] and [B] operations, and in FsDetView, query features are directly fed into RPN without aggregation of [A] operation.}

\label{fig:architecture}
\end{figure*}

\hj{Figure \ref{fig:architecture} shows the overall architecture illustrating two baselines \cite{fan2020fsod, xiao2020few} we used \nj{for FSOD}. As shown in the figure, there are two candidate \nj{locations} to apply the \nj{proposed} aggregation \nj{modules}. \nj{For different baselines \cite{fan2020fsod} and \cite{xiao2020few}, we designed similar aggregation methods composed of ISAM and QSAM and applied them at different locations as shown by [A] and [B] in Fig.~\ref{fig:architecture}}.} 

Specifically, the overall architecture is based on the two-stage Faster R-CNN framework \cite{ren2015faster}. Backbone network receives a query image and $K$ samples from the support set of the same class and outputs $K+1$ feature maps. \nj{The} \textit{Region Proposal Network} (RPN) proposes candidate boxes from the query feature map. Here, the input for \nj{the} RPN depends on the baseline. In FewX \cite{fan2020fsod}, the aggregated query feature map \hj{is} fed into RPN \nj{([A] in Fig.~\ref{fig:architecture})}. On the other hand, in FsDetView \cite{xiao2020few}, the query feature map is fed directly into the RPN without aggregation. In both baselines, $N_{roi}$ query RoI features are aggregated with $K$ support RoI features. Then, \textit{RoI Head} outputs box offsets and class confidences. 

\begin{figure*}
\centering
\includegraphics[width=16.5cm]{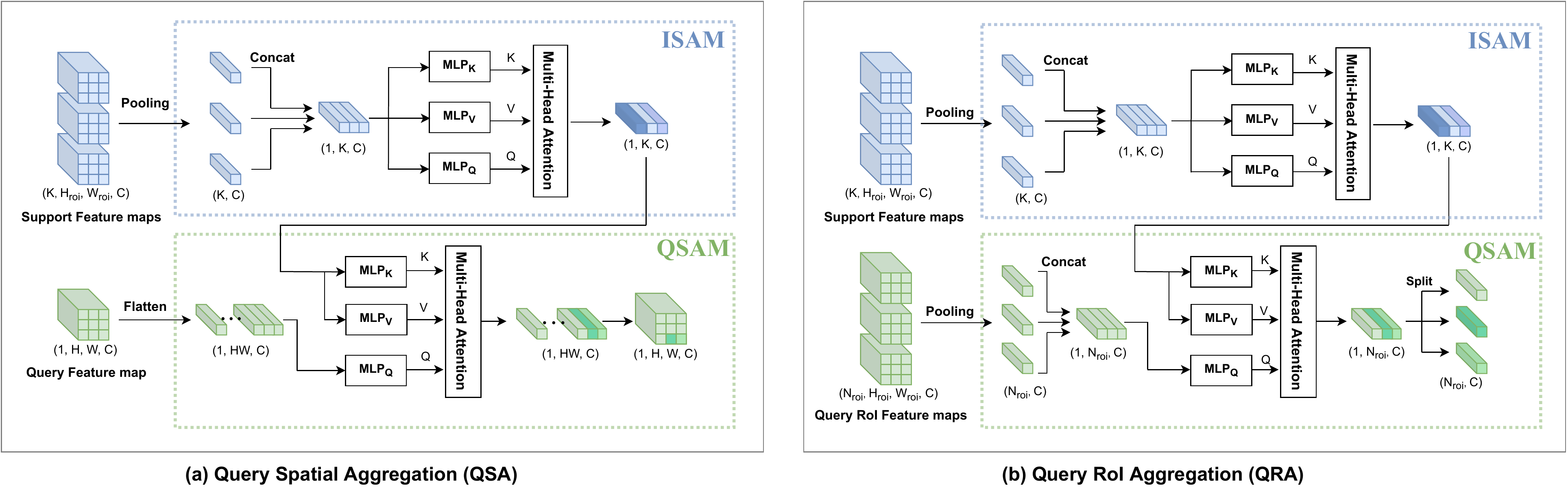}
\vspace{-0.1cm}
\caption{Two types of aggregation procedures (3-shot example) of the query feature maps and $K$ support feature maps. (a) is for spatial attention to the query feature map and (b) is for aggregation of $N_{roi}$ query feature maps and $K$ support feature maps.}
\label{fig:aggregation}
\end{figure*}

Note that both baselines \cite{fan2020fsod, xiao2020few} generate a single prototype per class by averaging the support feature maps before aggregation \nj{at} [A] for FewX and \nj{at [B]} for FsDetView. Unlike the two baselines, our method performs aggregation by treating the support features as an individual prototype, i.e., per-sample prototype. The detailed per-sample aggregation procedures of ISAM and QSAM are introduced in Section \ref{subsec:Aggregation} and Figure \ref{fig:aggregation}.

\subsection{Aggregation Module}
\label{subsec:Aggregation}

We propose a novel approach to aggregate the query and support features by treating each support feature as an individual prototype rather than generating a single prototype per category. Figure \ref{fig:aggregation} shows the aggregation methods according to the type of query feature map: (a) is for spatially boosting the prominence of the areas in the query feature map similar to the support RoI feature vectors, and (b) is to strengthen further the query RoI feature similar to support RoI feature vectors. Query Spatial Aggregation (a) is applied to Fig. \ref{fig:architecture} [A], and Query RoI Aggregation (b) is applied to Fig. \ref{fig:architecture} [B].
As shown in the figure, \nj{both of our aggregation methods are composed of two stages of} \textit{Intra-Support Attention Module} (ISAM) to refine the support features through the self-attention mechanism before aggregation and \textit{Query-Support Attention Module} (QSAM) that aggregates the queries with the supports.

\noindent{{\bf Intra-Support Attention Module}} aims to refine \nj{each individual support feature vector with the help of other support feature vectors}. Because some support feature vectors may be too far from other vectors due to the diverse nature of the support images, it can lead to performance degradation.
For example, an upright instance of a person can represent the characteristics of the \emph{person} class well, while an instance of a person doing a handstand cannot. 
Therefore, \nj{attention mechanism \cite{vaswani2017attention} shown in Eq. (\ref{equ:Attention}) is utilized to enhance the shared information of the support features. Especially, following  \cite{vaswani2017attention}, we use multi-head attention:}
\begin{equation} \label{equ:Attention}
\begin{split}
\mathtt{Q}_{attn}=Attention(\mathtt{Q}, \mathtt{K}, \mathtt{V})=softmax(\frac{\mathtt{Q} \mathtt{K}^{T}}{\sqrt{d}}) \mathtt{V} \\
\mathtt{Q}_{attn}, \mathtt{Q} \in \mathbb{R}^{q\times d}, \quad \mathtt{K}, \mathtt{V} \in \mathbb{R}^{v\times d} \quad
\end{split}
\end{equation}
where $d$ is the length of a feature vector. $q$ and $v$ are the number of \nj{queries and keys respectively}. Here, all of $\mathtt{Q}$, $\mathtt{K}$ and $\mathtt{V}$ are assigned to the support feature vectors, i.e., the number of support samples $q=v=K$. Then, the support feature vectors can pay attention to one another so that the inherent characteristics of the data can be refined.

For implementation, ISAM is designed as the encoder of shallow \emph{Transformer} \cite{vaswani2017attention, wang2019learning} composed of \emph{multi-head attention network} consisting of the process of Eq. (\ref{equ:Attention}) and \emph{multi-layer perceptron} with \emph{layer normalization} \cite{ba2016layer}.

\vspace{+0.15cm}
\noindent{{\bf Query-Support Attention Module}} aggregates the query feature and support feature maps through attention mechanism \nj{shown in} Eq.(\ref{equ:Attention}). Here, $\mathtt{K}$ and $\mathtt{V}$ are assigned to the support feature vectors, and $\mathtt{Q}$ is assigned to the query feature vectors. The query feature vectors are generated in one of two processes: by flattening the query feature map in (a), i.e.,
\hj{$q$=$HW$}, 
or by concatenating query RoI feature vectors in (b), i.e., $q$=$N_{roi}$. In other words, aggregation the query features and the support features are performed by dot-producting each query feature vector with all of the support feature vectors. 

For implementation, QSAM is designed as the decoder of shallow \emph{Transformer} which is also composed of \emph{multi-head attention network} containing \emph{multi-layer perceptron} blocks with \emph{layer normalization}.

\subsection{Training and Inference}
\label{subsec:objective}

\noindent{{\bf Training}} \quad Our framework is trained by two phases. First, the network is trained with abundant labeled base data $D_b$ with base class $C_b$. At this phase, the trained classes are $C_b$, i.e., $C_{train} = C_b$.
Second, the network is finetuned with few-shot novel data $D_n$ of novel classes $C_n$. At this phase, the training is done on a balanced dataset composed of $K$-shot instances per class for both base data and novel data, i.e., $C_{train} = C_{b} \cup C_{n}$.
For both phases, the episodic training strategy is applied that each episode consists of $N$-way, $K$-shot support data and a query image. FewX \cite{fan2020fsod} are trained with 2-way, $K$-shot. Specifically, an episode consists of the following triplet: ($q_{c1}$, $s_{c1}$, $s_{c2}$) where class ${c1}$ and ${c2}$ are different classes sampled from $C_{train}$, and $q_{c1}$ indicates the query data containing instances of ${c1}$-class. $s_{c1}$ and $s_{c2}$ indicate the $2$-way, $K$-shot support data, i.e., $|s_{c1}| = |s_{c2}| = K$. And for FsDetView \cite{xiao2020few}, an episode consists of a query image and all class of support data, i.e., $|C_{train}|$-way, $K$-shot.

\vspace{+0.15cm}
\noindent{{\bf Objective function}} \quad The loss function of RPN's foreground proposal and RoI Head's detection outputs are Eq. (\ref{equ:loss}) where $\mathcal{L}_{\cdot,loc}$ is the bounding box regression loss calculated as the smooth $\mathcal{L}_{1}$ loss, and $\mathcal{L}_{\cdot,cls}$ is the classification loss calculated as the cross-entropy loss. Note that the output of FewX is binary classification whether the query RoI feature vectors match or not with the support RoI feature vectors, and FsDetView is multi-class classification with the softmax function. $\mathcal{L}_{meta}$ is the cross-entropy loss used in FsDetView like Meta R-CNN \cite{yan2019meta} for class features to be diverse for different classes.
\begin{equation}
    \mathcal{L}= \mathcal{L}_{rpn,loc} + \mathcal{L}_{rpn,cls} + \\ \mathcal{L}_{det,loc} + \mathcal{L}_{det,cls} + \mathcal{L}_{meta}
    \label{equ:loss}
\end{equation}

\vspace{+0.15cm}
\noindent
{\bf Inference} \quad The few-shot samples used in finetuning are used as the support data at the inference time. Therefore, all the support data \nj{are} passed into the backbone network and ISAM once, and the output support feature vectors of ISAM are stored for repeated use as multiple prototypes.

\begin{table*}
\begin{center}
\begin{tabular}{|l|c|c|c|c|c|}
\hline
\multirow{2}{*}{method} &  \multicolumn{5}{c|}{Average precision at IoU=0.5}      \\ \cline{2-6} 
   &  K=1 & K=2 & K=3 & K=5 & K=10  \\
\hline\hline
FsDetView  & 23.8 $\pm$ 6.0 & 35.9 $\pm$ 6.1 & 42.1 $\pm$ 4.3 & 48.7 $\pm$ 3.4 & 56.9 $\pm$ 2.9 \\
FsDetView+ISAM & 24.0 $\pm$ 6.1 & 35.6 $\pm$ 5.2 & 44.0 $\pm$ 4.6 & 50.0 $\pm$ 4.0 & 57.9 $\pm$ 3.1 \\
FsDetView+QSAM & 23.9 $\pm$ 6.9 & 35.9 $\pm$ 5.5 & 43.9 $\pm$ 4.9 & 50.5 $\pm$ 3.9 & 58.1 $\pm$ 3.0 \\
FsDetView+ISAM+QSAM & \textbf{24.3 $\pm$ 6.2} & \textbf{36.5 $\pm$ 5.3} & \textbf{44.9 $\pm$ 4.3} & \textbf{52.0 $\pm$ 3.8} & \textbf{59.2 $\pm$ 2.6} \\ \hline
\end{tabular}
\end{center}
\caption{Ablation study on Novel set 1 of VOC07 test dataset. \nj{At base training, 3 support images per class are used}.} \vspace{+0.2cm}
\label{table:voc_abl_1_component}
\end{table*}

\begin{table*}
\begin{center}
\begin{tabular}{|l|c|c|c|c|c|c|}
\hline
\multirow{2}{*}{method} & base train & \multicolumn{5}{c|}{Average precision at IoU=0.5}      \\ \cline{3-7} 
   & K & K=1 & K=2 & K=3 & K=5 & K=10  \\
\hline\hline
FsDetView  & 1 & 23.8 $\pm$ 6.5 &  35.3 $\pm$ 6.0 & 41.8 $\pm$ 4.8 & 48.3 $\pm$ 3.6 & 56.5 $\pm$ 2.9 \\
FsDetView  & 3 & 23.8 $\pm$ 6.0 & 35.9 $\pm$ 6.1 & 42.1 $\pm$ 4.3 & 48.7 $\pm$ 3.4 & 56.9 $\pm$ 2.9 \\
FsDetView  & 10 & 23.2 $\pm$ 5.7 & 34.6 $\pm$ 5.9 & 41.5 $\pm$ 4.5 & 48.8 $\pm$ 3.4 & 57.1 $\pm$ 2.9 \\ \hline
FsDetView+ISAM+QSAM & 1 & \textbf{24.7} $\pm$ 6.5 & 35.9 $\pm$ 5.4 & 42.2 $\pm$ 4.4 & 50.4 $\pm$ 4.0 & 57.2 $\pm$ 3.0 \\ 
FsDetView+ISAM+QSAM & 3 & 24.3 $\pm$ 6.2 & \textbf{36.5 $\pm$ 5.3} & \textbf{44.9 $\pm$ 4.3} & \textbf{52.0 $\pm$ 3.8} & 59.2 $\pm$ 2.6 \\ 
FsDetView+ISAM+QSAM & 10 & 24.1 $\pm$ 5.9 & 36.2 $\pm$ 5.1 & 44.1 $\pm$ 4.2 & 51.8 $\pm$ 3.7 & \textbf{59.8 $\pm$ 2.2} \\ \hline
\end{tabular}
\end{center}
\caption{Ablation study with changes of $K$ \nj{during} base training on Novel set 1 of VOC07 test dataset.}
\label{table:voc_abl_2_num_shot_K}
\end{table*}

\section{Experiments}
\label{sec:experiments}

\subsection{Dataset}
\label{subsec:data}
We evaluate our method on PASCAL VOC \cite{pascal-voc-2007, pascal-voc-2012} and MS COCO \cite{lin2015microsoft} benchmarks. We follow the experimental setup of previous works \cite{yan2019meta,kang2019few, wang2020frustratingly, xiao2020few}. For VOC experiments, our network is trained using PASCAL VOC 07+12 trainval dataset and tested on VOC 07 test dataset. The 20 classes are divided into 15 base classes and 5 novel classes and evaluated with three different splits. The number of shots is set to $K \in \{1,2,3,5,10\}$. For 10-shot and 30-shot experiments on MS COCO, of the total 80 classes of COCO, \nj{the} 20 classes overlapping with those of VOC are set as novel classes. As in TFA \cite{wang2020frustratingly}, it is assumed that $K$-shot base data can be used when finetuning with $K$-shot novel data. Because the performance can vary depending on the few-shot sample configuration, we distinguished between the experiments in fixed support samples and the experiments in which random sampling is performed \nj{multiple times}.

\subsection{Implementation detail}
\label{subsec:imple}
The query images are resized to short size 600 and the maximum long side is set to be 1000. The support images are resized to 320x320 for FewX and 224x224 for FsDetView. We use Resnet-101 \cite{he2016deep} for the VOC experiment and Resnet-50 for the COCO experiment as the weight-shared backbone network except \nj{the} first convolution layer of FsDetView. Because \nj{input support data for FsDetView have 4 channels that consists of 3 rgb channels and 1 channel for the binary mask of ground truth bounding box,} FsDetView has a 3-channel convolution layer for the query image and a 4-channel convolution layer for the support data. The backbone networks are pretrained on Imagenet-1k \cite{ILSVRC15}. The learnable parameters of batchnorm \cite{ioffe2015batch} trained on imagenet 1k are frozen at both base training and finetuning \nj{for} both baselines. Learning schedulings are the same \nj{for both} FewX and FsDetView including the optimizer, learning rate, batch size and training/finetuning iterations. Unless otherwise noted, the numbers of support samples $K$ during base training of FewX, FewX+Ours, FsDetView and FsDetView+Ours are set to 10, 10, 1 and 3\footnotemark, respectively.  
\footnotetext{The $K$ during base training of FewX and FsDetView without ours followed the official code.}

ISAM and QSAM are implemented by utilizing the encoder and the decoder of the \textit{Transformer} \cite{vaswani2017attention}, respectively. Both are set to have 2-heads, 2-layers with \textit{layer normalization}, and \textit{ReLU} is used as an activation function. The dropout \cite{srivastava2014dropout} rate of Transformer is 0.1. And the hidden dimensions of the Transformer are set to 256.

\subsection{Analysis for ISAM and QSAM}
\label{subsec:ablation}
We applied our method to FsDetView \cite{xiao2020few} \nj{and analyzed how the performance changes on PASCAL VOC by two experiments}: ablation study of ISAM and QSAM (Table \ref{table:voc_abl_1_component}), and the effect of \nj{the number of per-class samples $K$} during base training (Table \ref{table:voc_abl_2_num_shot_K}). All experiments in Table \ref{table:voc_abl_1_component} and Table \ref{table:voc_abl_2_num_shot_K} were conducted 30 times to calculate the means and standard deviations.

\vspace{+0.15cm}
\noindent{}{\bf Ablation study}\quad 
Table \ref{table:voc_abl_1_component} shows the average precision at IoU=0.50 (Intersection over Union) of the novel classes with or without ISAM and QSAM on PASCAL VOC07 test dataset. It can be seen that it is effective to refine support features by paying attention to \nj{other support features} through ISAM. In addition, it can be seen that aggregation with the support features as they are through QSAM is more effective than generating a single class prototype by averaging support features. The table shows that both modules helped improve detection performance, and both processes work better as the number of shots $K$ increased.

\vspace{+0.15cm}
\noindent{}{\bf The number of \nj{shots} $K$ during base training}\quad Table \ref{table:voc_abl_2_num_shot_K} shows the AP50 results of the novel classes on PASCAL VOC07 test dataset according to the change of $K$ during base training. Our performances are generally higher, but the more similar the $K$ \nj{for} base training and the $K$ \nj{for} finetuning, the higher the performance. When FsDetView is trained by base data with \nj{ours} at $K=1$, ISAM did not learn how to pay attention to support features, and QSAM did not learn how to aggregate multiple prototypes together. Even if ISAM and QSAM learn \nj{their} roles when finetuning, the performances were lower than those of $K={3}$ or $10$ when base training. In addition, the higher the number of shots during base training, the lower the standard deviations.

\begin{figure*}[!t]
\centering
\includegraphics[width=17.4cm]{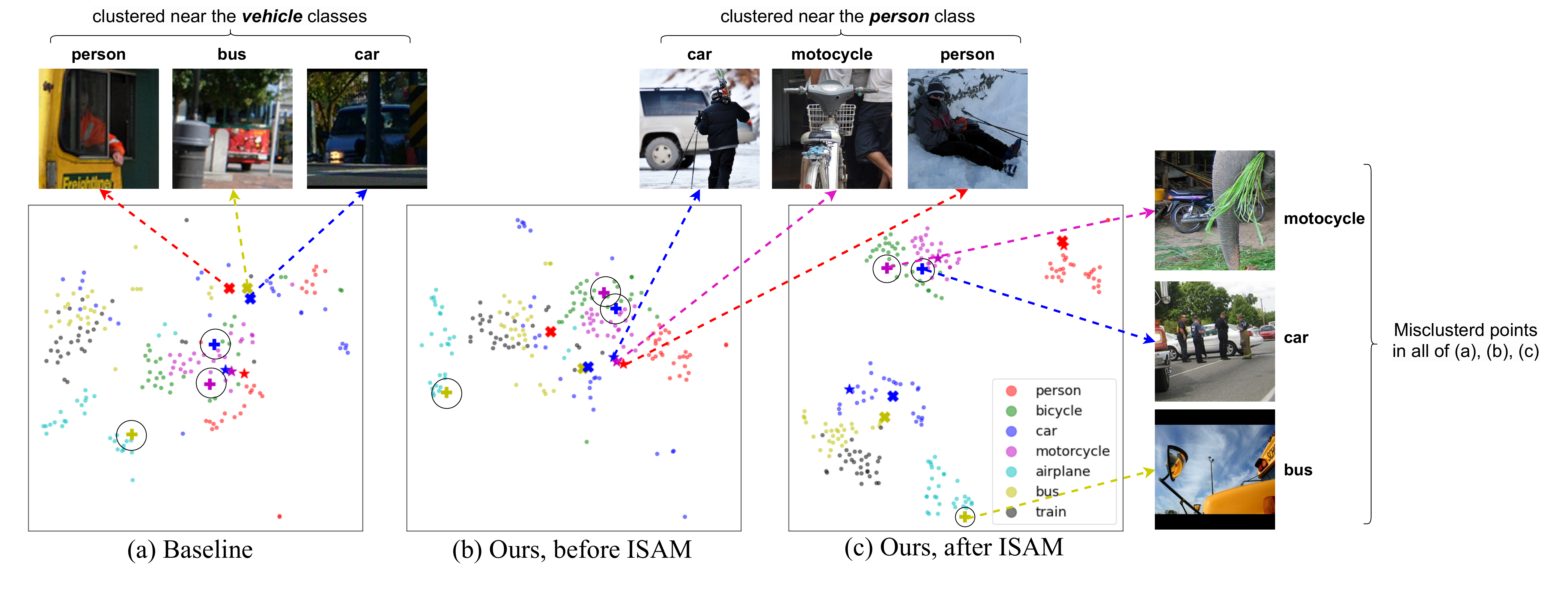}
\vspace{-0.9cm}
\caption{t-SNE visualization for 30-shot support feature vectors of MS COCO novel classes. (b) and (c) are support feature vectors before and after our ISAM, respectively. For visualization, some samples are marked with different markers (+,X,*), and only some classes are plotted. \nj{Some patches shown on the rightmost column, which are considered as misclusterd} in all of (a), (b), and (c), are circled.}
\label{fig:tsne_supp_attn}
\end{figure*}

\subsection{Clustering of support feature vectors}
\label{subsec:clustering}
\noindent{}{\bf t-SNE}\quad Our hypothesis is that collected natural images can be far from class prototypes. Hence, it is better that support features are refined into shared information through ISAM by paying attention to \nj{other} support features. Figure \ref{fig:tsne_supp_attn} is the t-SNE \cite{van2008visualizing} results of the novel classes on COCO 30-shot to verify the hypothesis. As in (a) and (b) of the figure, some features exist close to others \nj{despite} the categories are different. In \nj{these cases}, the data are noisy; for example, there are instances of several categories together in RGB images \nj{(a), (b)}. However, as can be seen in (c), ISAM makes clustering better for support features by paying attention to other support features. Some points are misclustered in all of (a), (b), and (c) when there are ambiguities in RGB images, such as multiple categories, occlusions, or partial appearances.

\vspace{+0.15cm}
\noindent{}{\bf Distance from class centroid} \quad We evaluated quantitatively whether each support feature vector, which we plotted on t-SNE, is actually closest to the corresponding class \nj{mean} (single prototype) calculated by 30-shots on the novel support data of COCO 30-shot. The accuracy was evaluated by measuring L1-distance with class \nj{means}. The accuracies of Baseline, before ISAM and after ISAM are 75.2 \%, 77.8\% and 97.8\%, respectively.

\subsection{Comparison with state-of-the-art}
\label{subsec:quant}

We applied our method to two baselines \cite{fan2020fsod, xiao2020few} and compared it with other methods on PASCAL VOC and COCO benchmarks. Note that the base models are trained by base data with $K$=10 for our method with FewX and $K$=3 for ours with FsDetView, as mentioned in implementation details (Sec. \ref{subsec:imple}). Both models are finetuned from each base model and are evaluated on novel classes.

\vspace{+0.15cm}
\noindent{}{\bf PASCAL VOC}\quad Table \ref{table:voc_novel} shows the AP50 results of the novel classes on PASCAL VOC07 test dataset. We evaluated Ours with FewX \cite{fan2020fsod} with the same few-shot samples as MetaYOLO \cite{kang2019few} and TFA \cite{wang2020frustratingly}, and there are significant performance improvements compared to the baseline. However, as shown in Table \ref{table:voc_abl_1_component}, the variance of performance is large. Therefore, we evaluated ours with FsDetView \cite{xiao2020few} by averaging 30 times of random samplings of few-shot samples. Likewise, significant performance improvements are found in FsDetView \cite{xiao2020few}.

\begin{figure}
    \centering
    \includegraphics[width=6.8cm]{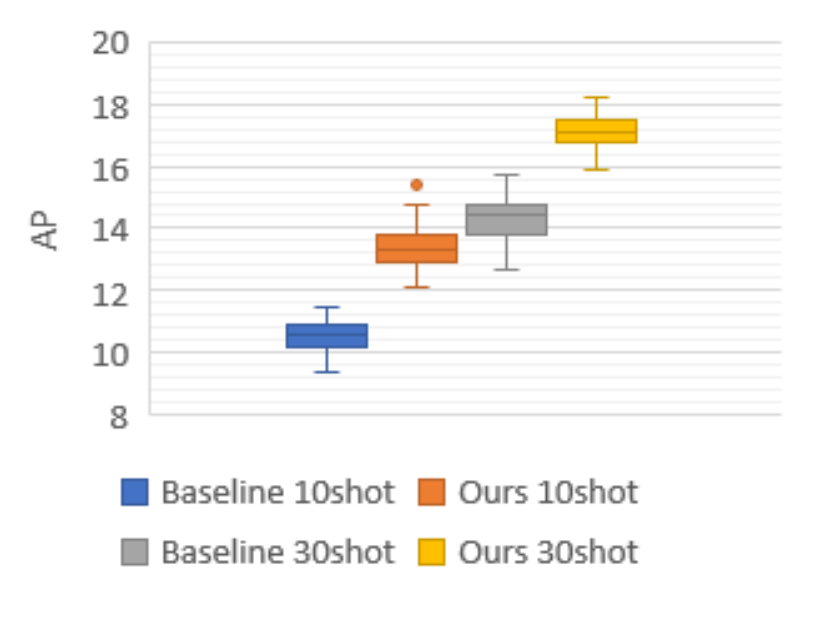}
    \vspace{-0.3cm}
    \caption{Box plot of Ours with FsDetView on MS COCO}
    \label{fig:coco_box_plot}
\end{figure}

\vspace{+0.15cm}
\noindent{}{\bf MS COCO}\quad Table \ref{table:coco_novel} summarizes the results for novel classes on MS COCO dataset, and we report the standard COCO metrics, average precision (AP) and average recall (AR). As shown in the table, our methods outperform the baselines in both cases of 10 shots and 30 shots. In addition, the box plot is visualized by repeating it 30 times in Fig. \ref{fig:coco_box_plot}. As shown in the figure, if the networks learned the same shot, it is confirmed that the min value of ours is higher than the max value of the baseline.

\begin{table*}
\begin{center}
\setlength\tabcolsep{1.8pt}
\begin{tabular}{|l|c|| c|c|c|c|c|| c|c|c|c|c|| c|c|c|c|c|}
\hline
\multirow{2}{*}{method} & \multirow{2}{*}{backbone} & \multicolumn{5}{c|}{Novel Set 1} & \multicolumn{5}{c|}{Novel set 2} & \multicolumn{5}{c|}{Novel set 3} \\ \cline{3-17} 
  &  &  1 & 2 & 3 & 5 & 10 & 1 & 2 & 3 & 5 & 10 & 1 & 2 & 3 & 5 & 10 \\
\hline\hline
    LSTD   \cite{chen2018lstd}     & \multirow{4}{*}{Darknet19 } & 8.2 & 11.0 & 12.4 & 29.1 & 38.5 & 11.4 & 3.8 & 5.0 & 15.7 & 31.0 & 12.6 & 8.5 & 15.0 & 27.3 & 36.3 \\    
    YOLOv2-ft \cite{kang2019few}  &  & 6.6 & 10.7 & 12.5 & 24.8 & 38.6 & 12.5 & 4.2 & 11.6 & 16.1 & 33.9 & 13.0 & 15.9 & 15.0 & 32.2 & 38.4 \\    
    MetaYOLO  \cite{kang2019few}  &  &14.8 & 15.5 & 26.7 & 33.9 & 47.2 & 15.7 &15.3 &22.7 &30.1 &40.5 &21.3 &25.6 &28.4&42.8 &45.9 \\    
    MetaDet  \cite{wang2019meta}   &  &17.1 & 19.1 & 28.9 & 35.0 & 48.8 & 18.2 &20.6 &25.9 &30.6 &41.5 &20.1 &22.3 &27.9& 41.9 &42.9 \\ \hline
    MetaDet  \cite{wang2019meta}    & VGG16 \cite{simonyan2014very} &18.9 & 20.6 & 30.2 & 36.8 & 49.6 & 21.8 &23.1 &27.8 &31.7 &43.0 &20.6 & 23.9 & 29.4 & 43.9 &44.1 \\
    NP-RepMet \cite{yang2020restoring}     & R101DCN-FPN    & 37.8 & 40.3 & 41.7 & 47.3 & 49.4 & 41.6 & 43.0 & 43.4 & 47.4 & 49.1 & 33.3 & 38.0 & 39.8 & 41.5  & 44.8 \\
    TFA w/ fc $\ast$ \cite{wang2020frustratingly} & R101FPN     & 36.8 & 29.1 & 43.6 & 55.7 & 57.0 & 18.2 & 29.0 & 33.4 & 35.5 & 39.0 & 27.7 & 33.6 & 42.5 &  48.7 & 50.2\\
    TFA w/ cos $\ast$ \cite{wang2020frustratingly} & R101FPN     & 39.8 & 36.1 & 44.7 & 55.7 & 56.0 & 23.5 & 26.9 & 34.1 & 35.1 & 39.1 & 30.8 & 34.8 & 42.8 &  49.5 & 49.8\\
    Meta R-CNN \cite{yan2019meta}     & R101      &19.9 & 25.5 & 35.0 & 45.7 & 51.5 & 10.4 &19.4 &29.6 &34.8 &45.4 &14.3 &18.2 &27.5&41.2 &48.1 \\
    
    FewX $\dagger$         & R101     & 29.8 & 35.5 & 36.3 & 48.4 & 53.6 & 22.2 & 28.9 & 25.2  & 31.2  & 39.7 & 24.3 & 29.9 & 34.4 & 47.1 & 50.4 \\
    \rowcolor{Gray}
    FewX+Ours         & R101       & 31.1 & 36.1 & 39.2 & 50.7 & 59.4 & 22.9 & 29.4 & 32.1 & 35.4 & 42.7 & 24.3 & 28.6 & 35.0 & 50.0 & 53.6 \\ 
    \hline \hline 
    TFA w/ cos $\ast$ $\flat$ \cite{wang2020frustratingly} & R101FPN     & \textcolor{red}{25.3}  & \textcolor{blue}{36.4} & 42.1 & 47.9 & 52.8 & 18.3 & \textcolor{red}{27.5} & 30.9 & 34.1 & 39.5 & 17.9 & 27.2 & 34.3 &  40.8 & 45.6\\
    FsDetView $\flat$ \cite{xiao2020few}  & R101       &24.2 & 35.3 & \textcolor{blue}{42.2} & \textcolor{blue}{49.1} & \textcolor{blue}{57.4} & \textcolor{red}{21.6} & 24.6 & \textcolor{blue}{31.9} & \textcolor{blue}{37.0} & \textcolor{blue}{45.7} & 21.2 &30.0 &\textcolor{blue}{37.2} & \textcolor{blue}{43.8} & \textcolor{blue}{49.6} \\
    FsDetView $\dagger$ $\flat$ & R101  & 23.8 & 35.3 & 41.8 & 48.3 & 56.5 &  19.4 & 26.4 & 30.3 & 36.6 & 44.6 & \textcolor{blue}{21.7} & \textcolor{blue}{31.3} & 34.2 & 40.6 & 47.9 \\
    \rowcolor{Gray}
    FsDetView+Ours $\flat$     & R101       & \textcolor{blue}{24.3} & \textcolor{red}{36.5} & \textcolor{red}{44.9} & \textcolor{red}{52.0} & \textcolor{red}{59.2} & \textcolor{blue}{20.5} & \textcolor{red}{27.5} & \textcolor{red}{33.1} & \textcolor{red}{40.9} & \textcolor{red}{47.1} & \textcolor{red}{22.4} & \textcolor{red}{33.0} & \textcolor{red}{37.8} & \textcolor{red}{43.9} & \textcolor{red}{51.5} \\
    \rowcolor{Gray}
\hline
\end{tabular}
\end{center}  
\vspace{-0.10cm}
\caption{AP50 on VOC2007 test dataset. The \nj{first four rows} are based on YOLOv2 \cite{redmon2017yolo9000}, and the rest are based on Faster R-CNN \cite{ren2015faster} with/without FPN \cite{lin2017feature} or DCN \cite{dai2017deformable}. \nj{Methods with $\ast$ marks are} based on finetuning and the others are based on meta-learning.  $\dagger$ \nj{indicates the re-implemented version} using \nj{the} official code. $\flat$ marks \nj{mean} multiple-run results. \textcolor{red}{Red}/\textcolor{blue}{Blue} texts indicate the first/second best on multiple-run results.}
\label{table:voc_novel}
\end{table*}

\begin{table*}
\begin{center}
\begin{tabular}{|c|c|c|c|c|c|c|c|c|c|}
\hline
\rowcolor{white}
\multirow{2}{*}{shots} & \multirow{2}{*}{method} & \multirow{2}{*}{backbone} & image & \multicolumn{3}{c|}{AP} & \multicolumn{3}{c|}{AR} \\ \cline{5-10} 
  &  &  & size & AP50:95 & AP50 & AP75 & 1 & 10 & 100 \\
\hline\hline
\multirow{11}{*}{10}& MetaYOLO \cite{kang2019few}    & Darknet19   & 416x416 & 5.6 & 12.3 & 4.6 & 10.1 & 14.3 & 14.4 \\
& MetaDet \cite{wang2019meta} & VGG16  & - & 7.1 & 14.6 & 6.1 & 11.9 & 15.1 & 15.5 \\
& TFA w/ fc $\ast$ \cite{wang2020frustratingly}  & R101FPN &short800 & 10.0 & 19.2 & 9.2 & - & - & - \\
& TFA w/ cos $\ast$ \cite{wang2020frustratingly}  & R101FPN &short800 & 10.0 & 19.1 & 9.3 & - & - & - \\
& Meta R-CNN \cite{yan2019meta} & R50  & short600 & 8.7 & 19.1 & 6.6 & 12.6 & 17.8 & 17.9 \\
& FewX $\dagger$  \cite{fan2020fsod}  & R50 & short600 & 11.9 & 23.7 & 10.6 & 19.1 & 26.2 & 26.3 \\
\rowcolor{Gray}
\cellcolor{white}& FewX+Ours & R50 & short600 & 13.0 & 24.7 & 12.1 & 19.3 & 27.7 & 27.8 \\ \cline{2-10}
&  TFA w/ fc $\ast$ $\flat$ \cite{wang2020frustratingly} & R101FPN & short800 & 9.1 & 17.3 & 8.8 & - & - & - \\
&  TFA w/ cos $\ast$ $\flat$ \cite{wang2020frustratingly} & R101FPN & short800 & 9.1 & 17.1 & 8.8 & - & - & - \\
& FsDetView $\flat$ \cite{xiao2020few}  & R50 & short600 & \textcolor{blue}{12.5} & \textcolor{blue}{27.3} & \textcolor{red}{9.8} & \textcolor{blue}{20.0} & \textcolor{blue}{25.5} & \textcolor{blue}{25.7} \\
& FsDetView $\dagger$  $\flat$  & R50 & short600 & 10.6 & 25.5 & 6.3 & 18.1 & 23.8 & 23.9 \\
\rowcolor{Gray}
\cellcolor{white} & FsDetView+Ours $\flat$ & R50 & short600 & \textcolor{red}{13.4} & \textcolor{red}{30.6} & \textcolor{blue}{9.1} & \textcolor{red}{20.7} & \textcolor{red}{26.7}  & \textcolor{red}{26.8} \\
\hline \hline

\multirow{11}{*}{30} &  MetaYOLO \cite{kang2019few}   & Darknet19 & 416x416 & 9.1 & 19.0 & 7.6 & 13.2 & 17.7 & 17.8 \\
& MetaDet \cite{wang2019meta}  & VGG16 & - & 11.3 & 21.7 & 8.1 & 14.5 & 18.9 & 19.2 \\
&  TFA w/ fc $\ast$ \cite{wang2020frustratingly} & R101FPN & short800 & 13.4 & 24.7 & 13.2 & - & - & - \\
&  TFA w/ cos $\ast$ \cite{wang2020frustratingly} & R101FPN & short800 & 13.7 & 24.9 & 13.4 & - & - & - \\
& Meta R-CNN \cite{yan2019meta} & R50  & short600 & 12.4 & 25.3 & 10.8 & 15.0 & 21.4 & 21.7 \\
&  FewX$\dagger$ \cite{fan2020fsod}   & R50 & short600 & 13.8 & 25.8 & 13.5 & 20.8 & 30.8 & 31.0 \\
\rowcolor{Gray}
\cellcolor{white} & FewX+Ours & R50 & short600 & 15.3 & 29.3 & 14.5 & 21.2 & 31.7 & 32.1 \\ \cline{2-10}
&  TFA w/ fc $\ast$ $\flat$ \cite{wang2020frustratingly} & R101FPN & short800 & 12.0 & 22.2 & 11.8 & - & - & - \\
&  TFA w/ cos $\ast$ $\flat$ \cite{wang2020frustratingly} & R101FPN & short800 & 12.1 & 22.0 & 12.0 & - & - & - \\
&  FsDetView $\flat$ \cite{xiao2020few}    & R50 & short600 & \textcolor{blue}{14.7}  & 30.6  & \textcolor{blue}{12.2} & \textcolor{blue}{22.0} & 28.2 & 28.4 \\
& FsDetView $\dagger$ $\flat$   & R50 & short600 & 14.3  & \textcolor{blue}{31.5}  & 10.6 & 21.9 & \textcolor{blue}{28.7} & \textcolor{blue}{28.8} \\
\rowcolor{Gray}
\cellcolor{white} &  FsDetView+Ours $\flat$   & R50 & short600 &  \textcolor{red}{17.1} & \textcolor{red}{35.2} & \textcolor{red}{14.7}  & \textcolor{red}{24.8}& \textcolor{red}{31.2} & \textcolor{red}{32.0} \\

\hline
\end{tabular}
\end{center} 
\vspace{-0.10cm}
\caption{AP and AR of novel claases on MS COCO minival. $\dagger$ is re-implemented using official code. $\ast$ marks method based on finetuning. $\flat$ marks multiple-run results. \textcolor{red}{Red}/\textcolor{blue}{Blue} texts indicate the first/second best on multiple-run results.}
\label{table:coco_novel}
\end{table*}

\section{Future work}
\label{sec:future work}
\noindent{}{\bf Aggregation with spatial information of support data}\quad As the support feature vectors were abstracted, spatial information of the vectors disappeared. Therefore, it is worth noting that the attention mechanism operates to reflect \textit{spatial information of support images}. In other words, when ISAM refines the support data, it is helpful to pay attention to each other without the support feature maps being pooled by global average pooling.
Similarly, when QSAM aggregates the query with the support, it is helpful to aggregate with spatial information of support feature maps rather than that of support feature vectors.

\vspace{+0.15cm}
\noindent{}{\bf Toward class scalable detector}\quad We evaluate Ours with FewX trained only on the base data of COCO dataset without finetuning on the novel data. The novel 20-class AP is 7.1 (+0.9 point increase compared to baseline) even though the novel data were not finetuned. Because the current framework based on metric-based meta-learning has the form of $P(box | query$ $image$, $support$ $images)$ . If the networks learn how to match given support images with the query image well, detection can be performed without finetuning the novel classes. If this characteristic is well utilized, it may be more advantageous to detect many unknown classes in the same domain or to perform it incrementally.

\section{Conclusion}
\label{sec:conclusion}
There are studies on the Few-Shot Object Detection framework based on meta-learning that detects instances of support category in a query image. Based on this framework, we propose the \textit{Intra-Support Attention Module} (ISAM) and \textit{Query-Support Attention Module} (QSAM) applicable to various methods. ISAM performs an attention mechanism between support features of the same class to refine the information that may be noisy, and QSAM aggregates the query features and the support features by per-sample prototypes, not a single prototype per class for using unabridged information of support data. Better feature maps for detecting unseen novel classes in $K$-shot support data are generated through these two modules. We demonstrate the effectiveness of the proposed modules in that the support feature vectors are clustered when collected support samples are somewhat far from the prototype. And the performances are improved when the attention vectors were refined and aggregated as per-sample prototypes.

\clearpage

{\small
\bibliographystyle{ieee_fullname}
\bibliography{egbib}
}

\end{document}